\documentclass[11pt,a4paper]{article}
\usepackage[hyperref]{emnlp2020}
\usepackage{times}
\usepackage{latexsym}

\usepackage{microtype}
\usepackage{url}
\usepackage[utf8]{inputenc}
\usepackage{color,soul}
\setul{0.45ex}{0.25ex}
\usepackage{graphicx}
\usepackage{subcaption}
\usepackage{amsmath}
\usepackage{booktabs}
\usepackage{multirow, makecell}
\usepackage{tikz}
\usepackage{graphicx}
\usepackage{array}
\usepackage{pgfplots}
\usepackage{algorithm}
\usepackage{textcomp}
\usepackage{xcolor}
\usepackage{times}
\usepackage{latexsym}
\usepackage{amsmath}
\usepackage{amsfonts}
\usepackage[noend]{algpseudocode}
\usepackage{lipsum}
\usepackage{multirow}
\usepackage{framed}
\usepackage{graphicx}
\usepackage{pifont}
\usepackage{longtable}
\usepackage{tikz}
\usepackage{booktabs}
\usepackage[all]{nowidow}
\usepackage{amssymb}
\usepackage{enumitem}
\usepackage{color,soul}
\usepackage{bm}
\usepackage{arydshln}
\usepackage{xspace}
\usepackage{siunitx}

\usepackage{titling}

\aclfinalcopy

\newcommand{\PreserveBackslash}[1]{\let\temp=\\#1\let\\=\temp}
\newcolumntype{C}[1]{>{\PreserveBackslash\centering}p{#1}}
\newcolumntype{R}[1]{>{\PreserveBackslash\raggedleft}p{#1}}
\newcolumntype{L}[1]{>{\PreserveBackslash\raggedright}p{#1}}
\newcommand{\bert}{BERT\xspace} 
\newcommand{\roberta}{RoBERTa\xspace} 

\definecolor{t_yellow}{HTML}{92921c}
\definecolor{t_green}{HTML}{3c8575}
\definecolor{t_blue}{HTML}{8692c6}
\definecolor{t_orange}{HTML}{cba058}
\definecolor{t_magenta}{HTML}{ae4388}
\newcommand{\trigger}[1]{\texttt{\textcolor{t_magenta}{#1}}}
\newcommand{\labtok}[1]{\texttt{\textcolor{t_yellow}{#1}}}
\newcommand{\prompt}[1]{\texttt{\textcolor{t_green}{#1}}}

\newcommand*\justify{%
  \fontdimen2\font=0.4em
  \fontdimen3\font=0.2em
  \fontdimen4\font=0.1em
  \fontdimen7\font=0.1em
  \hyphenchar\font=`\-
}

\newcommand{\methodname}{\textsc{AutoPrompt}\xspace}
\newcommand{\methodurl}{\url{http://ucinlp.github.io/autoprompt}\xspace}

\newif\ifcomments
\commentstrue
\ifcomments
    \providecommand{\yasaman}[2][]{{\protect\color{blue}{[Yasaman:\textbf{#1} #2]}}}
    \providecommand{\rob}[2][]{{\protect\color{red}{[Rob:\textbf{#1} #2]}}}
    \providecommand{\eric}[2][]{{\protect\color{magenta}{[Eric:\textbf{#1} #2]}}}
    \providecommand{\taylor}[2][]{{\protect\color{orange}{[Taylor:\textbf{#1} #2]}}}
    \providecommand{\sameer}[2][]{{\protect\color{violet}{[Sameer:\textbf{#1} #2]}}}
\else
    \providecommand{\yasaman}[2][]{}
    \providecommand{\rob}[2][]{}
    \providecommand{\eric}[2][]{}
    \providecommand{\taylor}[2][]{}
    \providecommand{\sameer}[2][]{}
\fi

\title{
\methodname: Eliciting Knowledge from Language Models
\\ with Automatically Generated Prompts
}

\author{
\bf Taylor Shin\thanks{~~First three authors contributed equally.}$\:\,^{{\diamondsuit}}$ \hspace{0.3cm} \bf Yasaman Razeghi$^{*\diamondsuit}$ \hspace{0.3cm} \bf Robert L. Logan IV$^{*\diamondsuit}$  \\ \bf Eric Wallace$^\spadesuit$ \hspace{0.3cm} \bf Sameer Singh$^\diamondsuit$ \\
$^\diamondsuit$University of California, Irvine \hspace{0.3cm} $^\spadesuit$University of California, Berkeley\\
\{\href{mailto:tshin1@uci.edu}{\tt tshin1},
\href{mailto:yrazeghi@uci.edu}{\tt yrazeghi}, \href{mailto:rlogan@uci.edu}{\tt rlogan},
\href{mailto:sameer@uci.edu}{\tt sameer}\}\href{mailto:sameer@uci.edu}{\tt @uci.edu}\\
\href{mailto:ericwallace@berkeley.edu}{\tt ericwallace@berkeley.edu}
}

\date{}

\begin{document}
\maketitle

\begin{abstract}
The remarkable success of pretrained language models has motivated the study of what kinds of knowledge these models learn during pretraining.
Reformulating tasks as fill-in-the-blanks problems (e.g., cloze tests) is a natural approach for gauging such knowledge, however, its usage is limited by the manual effort and guesswork required to write suitable prompts.
To address this,
we develop \methodname, an \emph{automated} method to create
prompts for a diverse set of
tasks, based on a gradient-guided search.
Using \methodname, we show that masked language models (MLMs) have an inherent capability to perform sentiment analysis and natural language inference without additional parameters or finetuning, sometimes achieving performance on par with recent state-of-the-art supervised models.
We also show that our prompts elicit more accurate factual knowledge from MLMs than the manually created prompts on the LAMA benchmark,
and that MLMs can be used as relation extractors 
more effectively than supervised relation extraction models.
These results demonstrate that automatically generated prompts are a viable parameter-free alternative to existing probing methods, and as pretrained LMs become more sophisticated and capable, potentially a replacement for finetuning.
\end{abstract}

\section{Introduction}

Pretrained language models (LMs) have had exceptional success when adapted to downstream tasks via \emph{finetuning}~\cite{PetersELMo2018, devlin2018BERT}.
Although it is clear that pretraining improves accuracy, it is difficult to determine whether the knowledge that finetuned LMs contain is learned during the \textit{pretraining} or the \textit{finetuning} process.
How can we directly evaluate the knowledge present in pretrained LMs, be it linguistic, factual, commonsense, or task-specific?

Numerous techniques have been proposed to elicit such knowledge by analyzing pretrained LMs' internal representations.
A common strategy is to use probing classifiers---shallow classifiers that predict certain attributes using an LMs' representations as features~\cite{conneau2018you,liu2019linguistic}.
However, probing classifiers require additional learned parameters and are thus susceptible to false positives; high probing accuracy is \textit{not} a sufficient condition to conclude that an LM contains a certain piece of knowledge~\cite{hewitt2019designing,voita2020information}.
Attention visualization, another common technique, has a similar failure mode: attention scores may be correlated with, but not caused by the underlying target knowledge, leading to criticism against their use as explanations~\cite{jain2019attention,wiegreffe2019attention}. Both probing and attention visualizations also struggle to evaluate knowledge that cannot be represented as simple token- or sequence-level classification tasks.

\begin{figure*}[tb]
    \centering
    \includegraphics{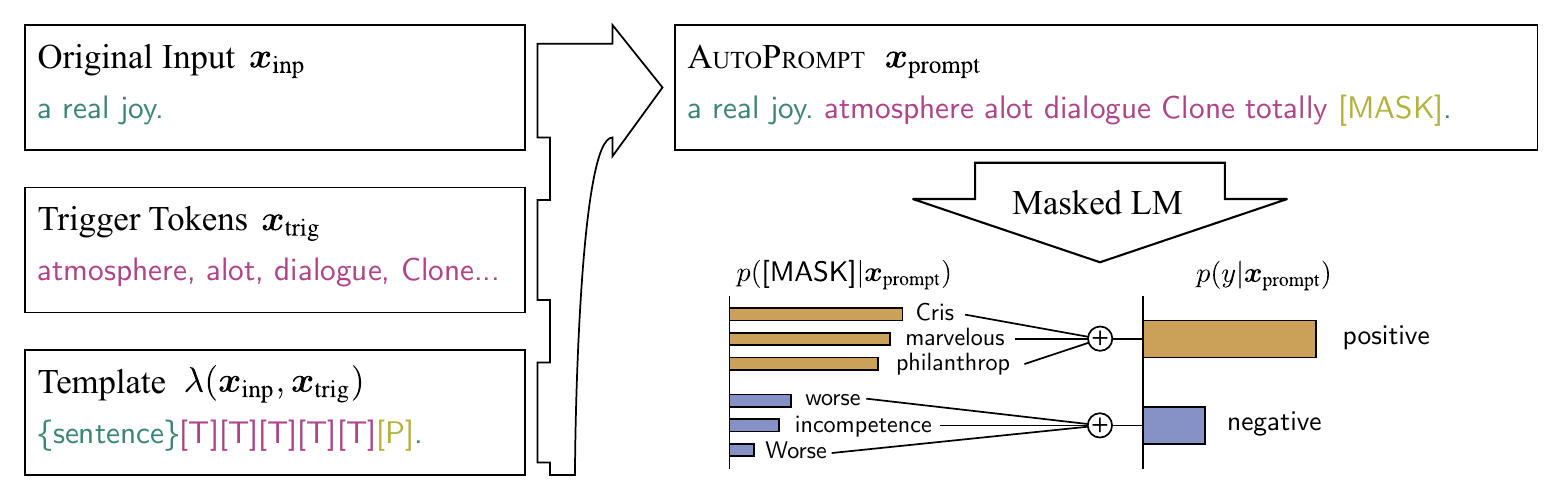}
    \caption{
        {\bf Illustration of \methodname} applied
        to probe a masked language model's (MLM's) ability to perform sentiment analysis.
        Each input, $\bm{x}_{\textrm{inp}}$, is placed into a natural language prompt, $\bm{x}_{\textrm{prompt}}$, which contains a single {\tt[MASK]} token.
        The prompt is created using a template, $\lambda$, which combines the original input with a set of trigger tokens, $\bm{x}_{\textrm{trig}}$.
        The trigger tokens are shared across all inputs and determined using a gradient-based search (Section~\ref{sec:prompt-construction}).
        Probabilities for each class label, $y$, are then obtained by marginalizing the MLM predictions, $p(\textrm{\tt[MASK]}|\bm{x}_{\textrm{prompt}})$, over sets of automatically detected label tokens (Section~\ref{sec:label}).
    }
    \label{fig:hook}
\end{figure*}

A more direct approach for eliciting knowledge from these models, since they are language models after all, is \emph{prompting}, i.e. converting tasks into a language model format.
For example, \citet{radford2019gpt2} frame summarization as a language modeling task by appending ``TL;DR:'' to the end of an article and then generating from an LM.
Similarly, \citet{petroni2019language} manually reformulate a knowledge base completion task as a cloze test (i.e., a fill-in-the-blank problem).
Compared to existing model analysis methods, prompting is non-invasive: it does not introduce large amounts of additional parameters or require direct inspection of a model's representations.
Thus prompting provides a lower bound on what the model ``knows'', and is therefore a more useful analysis tool.
However, prompting unfortunately requires manually crafting the context to feed into the model.
Not only is this time consuming and non-intuitive for many tasks (e.g., textual entailment), more importantly, models are highly sensitive to this context: improperly-constructed contexts cause artificially low performance~\cite{jiang2019can}. 
Overcoming the need to manually specify prompts would make prompting a more widely useful analysis tool.

In this paper, we introduce \methodname---an \emph{automated} method for generating prompts for any task,
illustrated in Figure~\ref{fig:hook}.
Given a task, e.g., sentiment analysis, \methodname creates a prompt by combining the original task inputs (e.g. reviews) with a collection of \emph{trigger tokens} according to a template.
The same set of trigger tokens is used for all inputs, and is learned using a variant of the gradient-based search strategy proposed in \citet{Wallace2019Triggers}.
The LM predictions for the prompt are converted to class probabilities by marginalizing over a set of associated label tokens, which can either be learned or specified ahead of time, enabling the LM to be evaluated the same as one would any other classifier.

We validate the effectiveness of \methodname in numerous experiments.
First, we use \methodname to construct prompts that test pretrained masked language models (MLMs) on sentiment analysis and natural language inference (NLI).
Our tests reveal that, without any finetuning, MLMs perform well on both of these tasks---a properly-prompted \roberta achieves 91\% accuracy on SST-2 (better than a finetuned ELMo model~\cite{PetersELMo2018}), and 69\% accuracy
on a balanced variant of the SICK-E dataset \cite{marelli2014sick}.
Next, we apply \methodname to the fact retrieval tasks of LAMA~\cite{petroni2019language}, where we are able to construct prompts that more effectively elicit MLM's factual knowledge than existing prompts generated using manual and corpus-mining methods.
Concretely, we achieve 43.3\% precision-at-1, compared to the current best single-prompt result of 34.1\%~\cite{jiang2019can}.
We also introduce a variant of this task, similar to relation extraction (RE), that tests whether MLMs can extract knowledge from a given piece of text.
We show that MLMs can actually \emph{outperform} existing RE models when context sentences with real facts are provided, however, they struggle when context sentences are artificially falsified.

Finally, although the goal of \methodname is to analyze models, we find that it provides certain practical advantages over finetuning. First, \methodname achieves higher average- and worst-case accuracy than finetuning in low-data regimes. Moreover, unlike finetuning, prompting LMs does not require large amounts of disk space to store model checkpoints; once a prompt is found, it can be used on off-the-shelf pretrained LMs. This is beneficial when serving models for multiple tasks.
\section{Overview of \methodname}

A natural way to elicit knowledge from pretrained LMs is to pose tasks as fill-in-the-blank problems.
However, writing prompts is not only time consuming, but it is not clear that the same phrasing will be effective for every model, nor is it clear what criteria determine whether a particular phrasing the \emph{best} to elicit the desired information.
In light of this, we introduce \methodname, a method that constructs customized prompts for a specific task and MLM of interest, to cause the MLMs to produce the desired knowledge.\footnote{Although we focus only on MLMs in this work, our method is trivially extendable to autoregressive LMs. The only adjustment is that the predict token must occur at the end of the prompt.}
An illustration of \methodname is provided in Figure~\ref{fig:hook}.
The prompt is constructed by taking the original task inputs---a collection of one or more sequences of tokens (e.g., the review in Figure \ref{fig:hook})---and mapping them to a sequence of tokens
using a template.
In the following sections, we describe how \methodname uses labeled training data to construct prompts, and how it uses the output of the MLM as a prediction for the task.

\subsection{Background and Notation}\label{sec:problem-setup}

For the purpose of prompt construction, we distinguish the original task inputs $\bm{x}_{\textrm{inp}}$ (e.g., the review in Figure \ref{fig:hook}, ``\prompt{a real joy.}")
from the prompt $\bm{x}_{\textrm{prompt}}$ (e.g., ``\prompt{a real joy.} \trigger{atmosphere alot dialogue Clone totally} \labtok{[MASK]}\prompt{.}") that is fed into the MLM.
The mapping from $\bm{x}_{\textrm{inp}}$ to $\bm{x}_{\textrm{prompt}}$ is performed using a template, $\lambda$.
This template defines where each input sequence will be placed in the prompt, as well as the placement of any additional tokens.
In particular, it must also define the placement of a special {\tt [MASK]} token for the MLM to fill in (denoted by [P] in the template to distinguish it from other {\tt [MASK]} tokens that might appear).
Feeding the prompt into the MLM produces a probability distribution $p(\textrm{\tt[MASK]}|\bm{x}_{\textrm{prompt}})$ describing which tokens most likely fill in the blank.

If class labels naturally correspond to tokens in the vocabulary (e.g., entity names in knowledge base completion tasks), this distribution may be readily interpreted as a distribution over class labels.
However, for tasks such as sentiment analysis, there may be a set of label tokens $\mathcal{V}_{y}$ that correspond to a particular label $y$.
For example, in Figure~\ref{fig:hook}, ``{\tt Cris}'', ``{\tt marvelous}'', and ``{\tt philanthrop}'' all indicate positive sentiment.
In this case, the class probability is obtained by marginalizing over the set of label tokens:
\begin{equation}\label{eqn:marginal-prob}
p(y|\bm{x}_{\textrm{prompt}}) = \sum_{w \in \mathcal{V}_{y}} p(\textrm{\tt[MASK]} = w | \bm{x}_{\textrm{prompt}})
\end{equation}

\subsection{Gradient-Based Prompt Search}
\label{sec:prompt-construction}

So far, we have shown how to reformulate a classification task as a language modeling task using prompts.
Here, we propose a method for \emph{automatic prompt construction} based on~\citet{Wallace2019Triggers}.
The idea is to add a number of ``trigger'' tokens that are shared across all prompts (denoted by  [T] in the example template in Figure \ref{fig:hook}).
These tokens are initialized to {\tt [MASK]} tokens, and then iteratively updated to maximize the label likelihood (Equation~\eqref{eqn:marginal-prob}) over batches of examples.

Formally, at each step, we compute a first-order approximation of the change in the log-likelihood that would be produced by swapping the $j$th trigger
token $x_{\textrm{trig}}^{(j)}$ with another token $w \in \mathcal{V}$.
Then we identify a candidate set $\mathcal{V}_{\textrm{cand}}$ of the top-$k$ tokens estimated to cause the greatest increase:
\begin{equation}
   \mathcal{V}_{\textrm{cand}} = \underset{w\in \mathcal{V}}{\text{top-}k} \left[\bm{w}_{\textrm{in}}^{T} \nabla \log p(y | \bm{x}_{\textrm{prompt}})\right]
\end{equation}
where $\bm{w}_{\textrm{in}}$ is the input embedding of $w$, and the gradient is taken with respect to the input embedding of $x_{\textrm{trig}}^{(j)}$.
Note that computing this candidate set is roughly as expensive as a single forward pass and backward pass of the model (the dot-products require the same amount of multiplications as computing the LM output projection). For each candidate in this set, we then re-evaluate Equation~\eqref{eqn:marginal-prob} on the updated prompt,
and retain the prompt with the highest probability in the next step---this requires $k$ forward passes of the model.
An example prompt produced by this method for the task of sentiment analysis is shown in Figure~\ref{fig:hook}.

\subsection{Automating Label Token Selection}\label{sec:label}

While in some settings the choice of label tokens is obvious (e.g., when class labels directly correspond to words in the vocabulary), it is less clear what label tokens are appropriate for problems involving more abstract class labels (e.g., NLI).   
In this section, we develop a general two-step approach to automate the selection of the sets of label tokens $\mathcal{V}_{y}$.
In the first step, we train a logistic classifier to predict the class label using the contextualized embedding of the {\tt [MASK]} token as input:
\begin{equation}\label{eqn:hidden}
\bm{h} = \textrm{Transformer}_{\textrm{enc}}(\tilde{\bm{x}})
\end{equation}
We write the output of this classifier as:
\begin{equation}\label{eqn:auto-label}
    p(y |\bm{h}^{(i)} ) \propto \exp(\bm{h}^{(i)} \cdot \bm{y} + \beta_{y})
\end{equation}
where $\bm{y}$ and $\beta_{y}$ are the learned weight and bias terms for the label $y$, and $i$ represents the index of the {\tt[MASK]} token.

In the second step, we substitute $\bm{h}^{(i)}$ with the MLM's output word embeddings $\bm{w}_{\mathrm{out}}$ to obtain a score $s(y, w) = p(y | \bm{w}_{\mathrm{out}})$.
Intuitively, because $\bm{w}_{out} \cdot \bm{h}$ and $\bm{y} \cdot \bm{h}$ are large for words and labels that are relevant to a particular context, $s_{w} \propto \exp (\bm{w}_{\textrm{out}} \cdot \bm{y} + \beta_y)$ should be large for words that are typically associated with a given label.
The sets of label tokens are then constructed from the $k$-highest scoring words:
\begin{equation}\label{eqn:label-tokens}
    \mathcal{V}_{y} =  \underset{w\in \mathcal{V}}{\text{top-}k} \left[ s(y, w) \right]
\end{equation}

\subsection{Relation to Other Prompting Methods}
\label{sec:related}

Our work fits into a body of work that probes language model's knowledge via prompts.
Previous works have used manually defined prompts to study an LM's ability to perform: commonsense reasoning~\cite{trinh2018simple, kwon2019masked, shwartz2020unsupervised}, question answering~\cite{lewis2019unsupervised},
fact recall~\cite{petroni2019language, jiang2019can, bouraoui2019inducing}, summarization~\cite{radford2019gpt2}, and other supervised tasks~\cite{brown2020language}.  \citet{schick2020exploiting} use manually constructed prompts in conjunction with semi-supervised learning for few-shot learning.
We instead \textit{automatically} create prompts for any task, which leads to higher accuracy and opens up new phenomena to analyze.

\subsection{Evaluation Setup}

In the following sections, we apply \methodname to probe BERT\textsubscript{BASE}\footnote{For brevity, we will omit subscripts in the model names.} 
(110M parameters) and  RoBERTa\textsubscript{LARGE}'s (355M parameters) knowledge of the following tasks: sentiment analysis, natural language inference (NLI), fact retrieval, and relation extraction.
We use the PyTorch implementations and pretrained weights provided by the \texttt{transformers} Python library~\cite{Wolf2019HuggingFacesTS}.
For sentiment analysis and NLI, we find label tokens using the logistic-regression-based heuristic described in Section~\ref{sec:label}.
For fact retrieval and relation extraction, we skip this step as the labels (entities) directly correspond to tokens in the vocabulary.
For all tasks, we perform the prompt search described in Section~\ref{sec:prompt-construction} for multiple iterations. In each iteration, we use a batch of training data to identify the candidate set $\mathcal{V}_{\textrm{cand}}$ of replacement trigger tokens. We then evaluate the label likelihoods of the updated prompts on a separate batch of data, and we retain the best trigger token in the next iteration of the search.
At the end of every iteration, we measure the label likelihood on withheld development data, and return the best prompt found during the entire search as the final output.
Performance is evaluated using the appropriate task-specific metrics---e.g., accuracy for sentiment analysis and NLI, and precision@$k$ for fact retrieval---on a separate withheld test set.

Our \methodname implementation is publicly available at \methodurl, and supports prompt generation for pretrained models in the HuggingFace {\tt transformers} library~\cite{Wolf2019HuggingFacesTS} on arbitrary datasets. 
\section{Sentiment Analysis}
\label{sec:sentiment}

Sentiment analysis is a fundamental task in NLP, both for natural language understanding research and real-world applications.
It is also difficult to probe the extent to which MLMs understand sentiment without finetuning.

\paragraph{Setup}
We apply our method to convert instances from the binary Stanford Sentiment Treebank ~\cite[SST-2]{socher2013recursive} into prompts, using the standard train/test splits.
We find label tokens using a prompt based on the template in Table~\ref{tab:example-templates}.
For our gradient-based prompt search, we perform a grid search over the following hyperparameters: 
$|\mathcal{V}_{cand}| \in \{10, 100\}$, $|\mathcal{V}_{y}| \in \{1, 3, 5\}$, $|\bm{x}_{\textrm{trig}}| \in [3,6]$.\footnote{Required 2 days to run with 8 NVIDIA 2080Ti GPUs.}
All prompts are initialized with the same template used to find the label set.

We also construct a prompt manually (before automated prompts are generated, to avoid bias) 
based on the intuition that SST-2 is comprised of movie reviews. We use
``\prompt{\{sentence\}} \trigger{this movie was} \labtok{[P]}\prompt{.}'' as the template, and use ``{\tt terrible}'' and ``{\tt fantastic}'' for the negative and positive label tokens, respectively.

\begin{table}[tb]
    \small
    \centering
    \begin{tabular}{l cc}
        \toprule
        {\bf Model } & {\bf Dev } & {\bf Test } \\
        \midrule
        BiLSTM & - & $82.8^{\ensuremath{\dagger}}$ \\
        BiLSTM + ELMo & - & $89.3^{\ensuremath{\dagger}}$ \\
        BERT (linear probing) & 85.2 & 83.4 \\
        BERT (finetuned) & - & $93.5^{\ensuremath{\dagger}}$ \\
        RoBERTa (linear probing) & 87.9 & 88.8 \\
        RoBERTa (finetuned) & - & $96.7^{\ensuremath{\dagger}}$ \\
        \midrule
        BERT (manual) & 63.2 & 63.2 \\
        BERT (\methodname) & 80.9 & 82.3 \\
        RoBERTa (manual) & 85.3 & 85.2 \\
        RoBERTa (\methodname) & 91.2 & 91.4 \\
        \bottomrule
    \end{tabular}
    \caption{
    \textbf{Sentiment Analysis}
        performance on the SST-2 test set of supervised classifiers (top) and fill-in-the-blank MLMs (bottom). Scores marked with~\ensuremath{\dagger} are from the GLUE leaderboard: \url{http://gluebenchmark.com/leaderboard}.
    }
    \vskip -5mm
    \label{tab:sst2-test}
\end{table}

\begin{figure}[th]
    \centering
    \begin{subfigure}[b]{\columnwidth}
        \centering
        \includegraphics[width=0.95\textwidth]{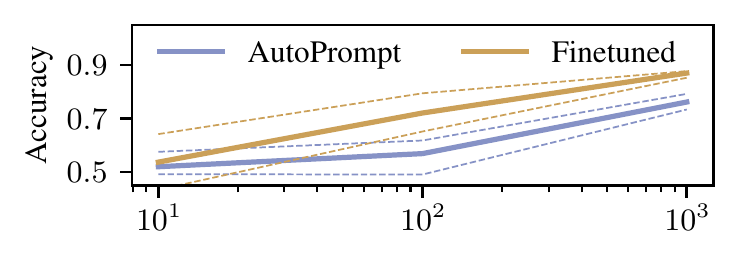}
        \caption{BERT on SST-2}
        \label{fig:training:sst2-bert}
    \end{subfigure}
    \qquad
    \begin{subfigure}[b]{\columnwidth}
        \centering
        \includegraphics[width=0.95\textwidth]{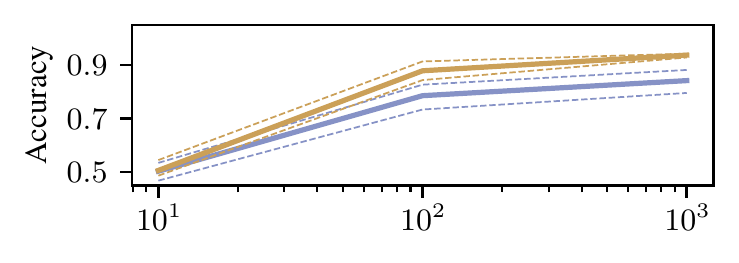}
        \caption{RoBERTa on SST-2}
        \label{fig:training:sst2-roberta}
    \end{subfigure}
    \begin{subfigure}[b]{\columnwidth}
        \centering
        \includegraphics[width=0.95\textwidth]{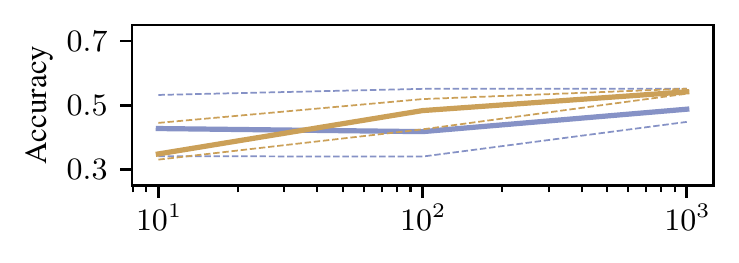}
        \caption{BERT on SICK-E}
        \label{fig:training:sicke-bert}
    \end{subfigure}
    \begin{subfigure}[b]{\columnwidth}
        \centering
        \includegraphics[width=0.95\textwidth]{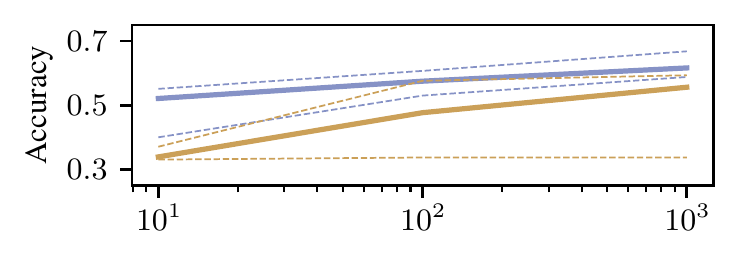}
        \caption{RoBERTa on SICK-E}
        \label{fig:training:sicke-roberta}
    \end{subfigure}
    \caption{
        \textbf{Effect of Training Data} on sentiment analysis and NLI for \methodname vs. finetuning. 
        X-axis is the number of data points used during training.
        Error bars plot the max. and min. accuracies observed over 10 independent runs.
        \emph{(revised since EMNLP version)}.
    }
    \label{fig:training-size}
\end{figure}

\paragraph{Results}
We show results in Table~\ref{tab:sst2-test}, along with reference scores from the GLUE~\cite{wang2018glue} SST-2 leaderboard, and scores for a linear probe trained over the elementwise average of the LM token representations.
Prompts generated by \methodname reveal that both BERT and RoBERTa have a strong knowledge of sentiment analysis:
without any finetuning, BERT performs comparably to a supervised BiLSTM, and RoBERTa achieves an accuracy on-par with finetuned BERT and ELMo models.
In addition, we observe that our automatically constructed prompts are more effective than manual prompts, and that they are difficult to construct using human intuition: the best template for RoBERTa is 
``{\tt \prompt{\{sentence\}} \trigger{atmosphere alot dialogue Clone totally} \labtok{[P]}\prompt{.}}''
We include results on the effect of the \methodname hyperparameters in Appendix~\ref{sec:sst2-hyper}.

\paragraph{Accuracy in Low-Data Settings}

Although the goal of \methodname is to probe a model's knowledge, we also find that it may be a viable alternative to finetuning in the low-data regime.
To show this, we measure the development set accuracy of \methodname prompts when using random subsets of 10, 100, and 1000 instances from the training data.
We run our prompt search with $|\bm{x}_{\textrm{trig}}|=10$, $|\mathcal{V}_{y}| = 3$, and $|\mathcal{V}_{\textrm{cand}}| = 10$.
We compare to the performance of BERT and RoBERTa finetuned on the same data.
For fair comparison between \methodname and finetuning, we use \citet{mosbach2020stability}'s recommended parameters for finetuning on small datasets: trained for 20 epochs, using AdamW~\cite{loshchilov2018decoupled} with bias correction and a learning rate that linearly increases to \num{2e-5} in the first 10\% of iterations, and linearly decreases to $0$ afterwards.
Experiments are repeated 10 times on random subsets of data (and seeds for the finetuned models).
Best-case, worst-case, and average performance are shown in Figure~\ref{fig:training-size}.
Note that results in the EMNLP version had a bug that has since been fixed.

We observe that while finetuning outperforms \methodname on sentiment analysis, \methodname can perform better than finetuning on NLI.
Notably, \methodname elicits better average performance from both BERT and RoBERTa given only 10 training examples.
Furthermore, results for RoBERTa are more stable across all sample sizes whereas finetuning can result in ``failed runs'' (consistent with~\citealt{dodge2020fine}).
This behavior in the low-data regime is an interesting phenomenon, and suggests that there are barriers that MLMs must surmount when they are converted to finetuned classifiers that are not encountered when the task is presented as masked language modeling. 

\section{Natural Language Inference}
\label{sec:nli}

\begin{table}[tb]
    \small
    \centering
    \setlength{\tabcolsep}{4pt}
    \begin{tabular}{lccc}
        \toprule
        \multirow{2}{*}{\bf Model } & \multicolumn{3}{c}{\bf SICK-E Datasets} \\
        & standard & 3-way & 2-way\\
        \midrule
        Majority  & 56.7 & 33.3  & 50.0\\
        BERT (finetuned)  & 86.7 & 84.0 & 95.6\\
        BERT (linear probing) & 68.0 & 49.5 & 91.9\\
        RoBERTa (linear probing) & 72.6 & 49.4 & 91.1 \\
        \midrule
        BERT (\methodname)   & 62.3 & 55.4 &  85.7\\
        RoBERTa (\methodname)    & 65.0  & 69.3  & 87.3\\
        \bottomrule
    \end{tabular}
    \caption{\textbf{Natural Language Inference} performance on the SICK-E test set and variants.
    (Top) Baseline classifiers. (Bottom) Fill-in-the-blank MLMs.
    }
    \label{table:NLI results}
\end{table}

\begin{table*}[tb]
    \small
    \centering
    \begin{tabular}{C{0.40in}L{1.35in}L{1.95in}L{1.9in}}
        \toprule
        {\textbf{Task}} & {\bf Prompt Template} & {\bf Prompt found by \methodname} & {\bf Label Tokens}\\
        \midrule

        \multirow{1}{*}{\shortstack[c]{Sentiment\\ Analysis}}
        &   \vspace{-6pt}
            {\tt \prompt{\{sentence\}} \trigger{[T]\ldots[T]}  \labtok{[P]}\prompt{.} } &
            \vspace{-6pt}
            {\tt \prompt{unflinchingly bleak and desperate} \trigger{Writing academicswhere overseas will appear} \labtok{[MASK]}\prompt{.}} &
            \footnotesize{{\bf pos}: partnership, extraordinary, \#\#bla \newline
            {\bf neg}: worse, persisted, unconstitutional}
            \\
        \midrule
        
        \multirow{1}{*}{\shortstack[c]{NLI}} & 
            \vspace{-5pt}
            {\tt \prompt{\{prem\}}\labtok{[P]}\trigger{[T]\ldots[T]}\prompt{\{hyp\}}}
        & 
        \vspace{-7pt}
        \tt{\justify{\prompt{Two dogs are wrestling and hugging} \labtok{[MASK]} \trigger{concretepathic workplace} \prompt{There is no dog wrestling and hugging}}}
        & 
        \footnotesize{{\bf con}: Nobody, nobody, nor \newline
        {\bf ent}:  \#\#found, \#\#ways, Agency \newline
        {\bf neu}: \#\#ponents, \#\#lary, \#\#uated }
       \\
        
        \midrule
        \multirow{1}{*}{\shortstack[c]{Fact \\ Retrieval}} & 
            {\footnotesize \textit{X plays Y music}} \newline
            {\tt \prompt{\{sub\}}\trigger{[T]\ldots[T]}\labtok{[P]}\prompt{.} }
            &
            \vspace{-7pt}
            {\tt \prompt{Hall Overton}
            {\color{t_magenta} fireplacemade antique son alto} {\color{t_yellow} [MASK]}\prompt{.}} & 
            \\

        \midrule
        \multirow{1}{*}{\shortstack[c]{Relation \\ Extraction}} & 
            {\footnotesize \textit{X is a Y by profession}} \newline
            {\tt \prompt{\{sent\}\{sub\}}\trigger{[T]\ldots[T]}\labtok{[P]}\prompt{.}}
            &
            \vspace{-8pt}
            {\tt {\color{t_green} Leonard Wood (born February 4, 1942) is a former Canadian politician.}\newline
            {\color{t_green} Leonard Wood}
            {\color{t_magenta} gymnasium brotherdicative himself another} {\color{t_yellow} [MASK]}\prompt{.}} &
            \\
        \bottomrule
    \end{tabular}
    \caption{\textbf{Example Prompts} by \methodname for each task. On the left, we show the prompt template, which combines the input, a number of trigger tokens \trigger{[T]}, and a prediction token \labtok{[P]}. For classification tasks (sentiment analysis and NLI), we make predictions by summing the model's probability for a number of automatically selected label tokens. For fact retrieval and relation extraction, we take the most likely token predicted by the model.}
    \label{tab:example-templates}
\end{table*}

To evaluate the \textit{semantic} understanding of MLMs, we experiment on Natural Language Inference (NLI).
NLI is crucial in many tasks such as reading comprehension and commonsense reasoning~\cite{bowman2015large}, and it is used as a common benchmark for language understanding.

\paragraph{Setup}
We use the entailment task from the SICK dataset~\cite[SICK-E]{marelli2014sick} which consists of around 10,000 pairs of human-annotated sentences labeled as entailment, contradiction, and neutral.
The standard dataset is biased toward the neutral class which represent $56.7$\% of instances.
We also experiment on an unbiased variant with 2-way classification of contradiction vs. entailment (\textit{2-way}), as well as an unbiased 3-way classification variant (\textit{3-way}). 
The template used for \methodname is provided in Table~\ref{tab:example-templates}.
We search over the following parameters: $|\mathcal{V}_{cand}| \in \{10, 50\}$, $|\mathcal{V}_{y}| \in \{1, 3, 5, 10\}$, $|\bm{x}_{\textrm{trig}}| \in [1,5]$, and choose the best prompt according to development set accuracy.

\paragraph{Results}
Table~\ref{table:NLI results} shows that \methodname considerably outperforms the majority baseline in all experiments.
For example, on the 2-way SICK-E dataset, \methodname is comparable to a supervised finetuned BERT.
We also test linear probes---linear classifiers trained on top of frozen MLM representations with average pooling ---and find \methodname has comparable or higher accuracy, despite linear probes being susceptible to false positives. 
Overall, these results demonstrate that both BERT and RoBERTa have some inherent knowledge of natural language inference.

We also examine the efficacy of \methodname in the low-data regime (using the same procedure as SST-2) on the unbiased 3-way SICK-E data. The results in Figure~\ref{fig:training-size} show that \methodname performs on par with finetuned BERT and significantly better than finetuned RoBERTa in low data settings.

\paragraph{MLMs Excel on Contradiction}
We find that the label tokens are more interpretable for \emph{contradiction} compared to \textit{entailment} or \textit{neutral} (examples in Table~\ref{tab:example-templates}).
We investigate if this hurts the model performance on entailment and neutral classes. 
We measure the precision for each label in the 3-way balanced SICK-E dataset.
BERT achieves $74.9$\%, $54.4$\%, and $36.8$\% precision for contradiction, entailment, and neutral cases, respectively, while RoBERTa obtains $84.9$\%, $65.1$\%, and $57.3$\%. 
These results suggest that \methodname may be more accurate for concepts that can be easily expressed using natural label tokens. 

\section{Fact Retrieval}
\label{sec:factual}

An important question is whether pretrained MLMs \emph{know} facts about real-world entities.
The LAMA dataset~\cite{petroni2019language} evaluates this using cloze tests that consist of \texttt{(sub, rel, obj)} triples, e.g. (\texttt{Obama}, \texttt{bornIn}, \texttt{Hawaii}), and \textit{manually} created prompts with missing objects, e.g. ``\prompt{Obama} \trigger{was born in} \labtok{[MASK]}\prompt{.}''.
LPAQA~\cite{jiang2019can} extends this idea by \textit{systematically} creating prompts that are generated by mining Wikipedia, paraphrasing, and crowdsourcing. 
In this section, we use the same cloze-style setup but \emph{automatically} generate prompts in order to better evaluate the factual knowledge of MLMs. We compare our approach against LAMA and LPAQA, which are explicitly designed for the task of fact retrieval.

\paragraph{Setup}
We reformulate fact retrieval by mapping \texttt{(sub,rel,obj)} triples to a prompt using the template ``\prompt{\{sub\}}\trigger{[T]\ldots[T]}\labtok{[P]}\prompt{.}'', where the trigger tokens are specific to the relation \texttt{rel} and the correct object \texttt{obj} is the label token.
We use the original test set from LAMA~\cite{petroni2019language}, henceforth \emph{Original}. To collect training data for \methodname, we gather at most 1000 facts for each of the 41 relations in LAMA from the T-REx dataset~\cite{DBLP:conf/lrec/ElSaharVRGHLS18}. For the relations that still have less than 1000 samples, we gather extra facts straight from Wikidata. We ensure that none of the T-REx triples are present in the test set, and we split the data 80-20 into train and development sets. Moreover, because the collected T-REx data is from a slightly different distribution than the LAMA test set, we also consider a separate evaluation where we split the T-REx triples into a 60-20-20 train/dev/test split and evaluate on the test set. This \emph{T-REx} dataset is used to measure the performance of our prompts when the train and test data is from the same distribution.

We use \methodname with 5 or 7 tokens, and select the search parameters using the T-REx development set.
We prevent proper nouns and tokens that appear as gold objects in the training data from being selected as trigger tokens. 
This is done to prevent \methodname from ``cheating'' by embedding common answers inside the prompt.
To evaluate, we observe the rank of the true object in label token distribution of the MLM, and use standard ranking metrics: mean reciprocal rank (MRR), precision-at-1 (P@1), and precision-at-10 (P@10).

\begin{table*}[tb]
\begin{subtable}{.55\linewidth}
    \small
    \centering
    \begin{tabular}{lrrrrrr}
    \toprule
    \multirow{3}{*}{\bf Prompt Type} & \multicolumn{3}{c}{\bf Original} & \multicolumn{3}{c}{\bf T-REx} \\
    
     \cmidrule(lr){2-4}
     \cmidrule(lr){5-7}
     & MRR & P@10 & P@1  & MRR & P@10 & P@1 \\
    \midrule
    LAMA & 40.27 & 59.49 & 31.10 & 35.79 & 54.29 & 26.38 \\
    LPAQA (Top1) & 43.57 & 62.03 & 34.10 & 39.86 & 57.27 & 31.16 \\
    \methodname 5 Tokens & 53.06 & 72.17 & 42.94 & 54.42 & 70.80 & 45.40 \\
    \methodname 7 Tokens & 53.89 & 73.93 & 43.34 & 54.89 & 72.02 & 45.57 \\
    \bottomrule
    \end{tabular}
\end{subtable}%
\hspace{.12\linewidth}%
\begin{subtable}{.30\linewidth}
    \small
    \centering
    \begin{tabular}{lccc}
    \toprule
    \bf Model & MRR & P@10 & P@1 \\
    \midrule
    BERT & 55.22 & 74.01 & 45.23 \\
    RoBERTa & 49.90 & 68.34 & 40.01 \\
    \bottomrule
    \end{tabular}
    \end{subtable}
    \caption{\textbf{Factual Retrieval: } On the left, we evaluate BERT on fact retrieval using the \emph{Original} LAMA dataset from \citet{petroni2019language}. For all three metrics (mean reciprocal rank, mean precision-at-10 (P@10), and mean precision-at-1(P@1)), \methodname significantly outperforms past prompting methods. We also report results on a \emph{T-REx} version of the data (see text for details). On the right, we compare BERT versus RoBERTa on a subset of the LAMA data using \methodname with 5 tokens.
    }
\label{tab:fact-ret}
\end{table*}

\paragraph{Results}
Table~\ref{tab:fact-ret} shows the performance of MLMs with different prompting methods, and we show qualitative examples in Table~\ref{tab:example-templates} and in Appendix~\ref{app:factual}.
Prompts generated using \methodname can extract factual knowledge from BERT more effectively than their manual and mined counterparts: we improve P@1 by up to 12 points.
Moreover, despite \methodname using only one prompt per relation, it still outperforms LPAQA's ensemble method (which averages predictions for up to 30 prompts) by approximately 4 points.
Using 7 trigger tokens achieves slightly higher scores than 5 trigger tokens, although the difference is not substantial. This indicates that our approach is stable to the choice of trigger length, which is consistent with our sentiment analysis results.
Overall, these results show that \methodname can retrieve facts more effectively than past prompting methods, thus demonstrating that \bert contains more factual knowledge than previously estimated.

\paragraph{Relation Breakdown}
We also provide a detailed breakdown of the prompts found by \citet{petroni2019language} and \methodname, and their associated accuracies in Appendix~\ref{app:factual}, Table~\ref{tab:fact-ret-rob}. 
Manual prompts are competitive when the prompt is \textit{easy} to specify, e.g., the prompt \emph{``was born in''} for the \textsc{place of birth} relation.
On the other hand, \methodname performs especially well for relations that are difficult to specify in a natural language prompt.
For example, \citet{petroni2019language}'s prompt for the \textsc{position played on team} relation is ``\prompt{\{sub\}} \trigger{plays in} \labtok{[MASK]} \trigger{position}'', which is not as specific as the relation requires.
Although the prompt from \methodname is not grammatical (\emph{``\prompt{\{sub\}} \trigger{ediatric striker ice baseman defensive} \labtok{\{obj\}}''}), it does contain tokens that are directly related to sports.

\paragraph{BERT outperforms \roberta} We finally directly compare BERT and RoBERTa. To do so, we subsample the LAMA test set to consist of examples where the object is a single token for both BERT and RoBERTa (\textit{Original-RoBERTa}).\footnote{The original dataset consists of examples where the object is a single token for BERT.}
\bert actually slightly outperforms RoBERTa, and we find that the prompts generated for RoBERTa tend to contain more irrelevant words (see Appendix~\ref{app:factual}, Table~\ref{tab:fact-ret-rob}).
For example, the prompt generated by RoBERTa for the \textsc{plays instrument} relation contains words such as ``Trump'' and symbols such as ``,'' (),'' for the \textsc{position played on team} relation. 
It is surprising that \roberta does not perform better than \bert, and it is worthy of investigating this further in future work.
Additionally, recall that prompting is a \emph{lower bound} on a model's knowledge: the lower relative performance does not mean that the model actually knows less.

\section{Relation Extraction}
\label{sec:ie}

Apart from evaluating whether MLMs \emph{know} facts, it is also important to evaluate whether they can \emph{extract knowledge} from text.
In this section, we use the task of relation extraction (RE)---to identify how entities are related in a given sentence---an important task in information extraction.
We create RE prompts in a similar fashion as fact retrieval: for a given triple \texttt{(subj,rel,obj)} and sentence that expresses this relation, we construct a prompt as ``\prompt{\{sent\}\{sub\}}\trigger{[T]\ldots[T]}\labtok{[P]}\prompt{.}'', where the trigger tokens are specific to the relation, and label token is the correct object \texttt{obj} (see Table~\ref{tab:example-templates} for an example).

\paragraph{Setup}
We use the T-Rex dataset for RE because each T-REx fact comes with context sentences that mention the subject and object surface forms.
We compare \methodname to LAMA and LPAQA (their prompts are still useful here), as well as a recent supervised relation extraction model~\cite{DBLP:conf/emnlp/SorokinG17} that was also used by \citet{petroni2019language}. To make the evaluation fair for the supervised RE model, we modify the standard RE evaluation. We give the model credit as long as it does not predict a different relation for the subject and object, i.e. we ignore the ``no relation'' prediction and all other relations. We also drop all sentences from evaluation for which the model's named entity extractor failed to identify the subject and the object as entities.
See Appendix~\ref{app:re} for further details. For the evaluation of all systems, we treat a prediction as correct if it is either the canonical version of the object (e.g., ``USA'') or the rendered surface form (e.g., ``American'') for \textit{any} of the context sentences in a given triple.

\paragraph{Results}
Table~\ref{tab:RE} shows the results for BERT and RoBERTa. 
MLMs can extract relational information \textit{more effectively} than the supervised RE model,
providing up to a 33\% increase on the task when using \methodname.
RoBERTa also outperforms the supervised RE model, although it is worse than BERT (likely for similar reasons as we outline in Section~\ref{sec:factual}).
For both BERT and RoBERTa, we notice that the trigger tokens consist of words related to their corresponding relations (see Appendix~\ref{app:rel}, Table~\ref{tab:RE-examples} for full list), e.g. RoBERTa selects {\trigger{``defy trademarks of namesake manufacturer''}} for relation \textsc{manufacturer/producer of product}.

\paragraph{Perturbed Sentence Evaluation}
A possible explanation for the strong results of MLMs in the RE setting is that they may \textit{already} know many of the relations. Thus, they may directly predict the objects instead of \textit{extracting} them.
To separate this effect, we synthetically perturb the relation extraction dataset by replacing each object in the test data with a random other object and making the same change to the prompt. For example, \emph{``Ryo Kase (born November 9, 1974 in \st{Yokohama}$\rightarrow$Yorkshire) is a Japanese actor''} where \texttt{Ryo Kase} is the subject, \texttt{Yokohama} is the original object, and \texttt{Yorkshire} is the new object.
We regenerate the prompts using the perturbed version of the data.

The accuracy of the RE model does not change significantly on the perturbed data (Table~\ref{tab:RE}), however, the accuracy of the MLMs decreases significantly.
This indicates that a significant portion of MLM accuracy comes from background information rather than relation extraction.
Nevertheless, our prompts for BERT outperform their LAMA and LPAQA counterparts, which provides further evidence that \methodname produces better probes.

\begin{table}[tb]
\small
    \centering
    \begin{tabular}{lrr}
    \toprule
    \bf Model & \bf Original & \bf Perturbed \\
    \midrule
    Supervised RE LSTM & 57.95 & 58.81\\
    BERT (LAMA) & 69.06 & 28.02 \\
    BERT (LPAQA) & 76.55 & 30.79 \\
    BERT (\methodname) & 90.73 & 56.43 \\
    RoBERTa (\methodname) & 60.33 & 28.95 \\
    \bottomrule
    \end{tabular}
    \caption{\textbf{Relation Extraction:} We use prompts to test pretrained MLMs on relation extraction. Compared to a state-of-the-art LSTM model from 2017, MLMs have higher mean precision-at-1 (P@1), especially when using prompts from \methodname. We also test models on sentences that have been edited to contain incorrect facts. The accuracy of MLMs drops significantly on these sentences, indicating that their high performance stems from their factual knowledge.}
    \label{tab:RE}
\end{table}
\section{Discussion}
\label{sec:discussion}

\paragraph{Prompting as an Alternative to Finetuning}

The goal of prompting a language model is to probe the knowledge that the model acquired from pretraining. Nevertheless, prompting has some practical advantages over finetuning for solving real-world tasks. First, as shown in Section~\ref{sec:sentiment}, prompts generated using \methodname can achieve higher accuracy than finetuning in the \textit{low-data regime}. 
Moreover, prompting has advantages over finetuning when trying to solve \textit{many different tasks} (e.g., the many users of the OpenAI GPT-3 API~\cite{brown2020language}). In particular, finetuning requires storing large language model checkpoints for each individual task, and drastically increases system cost and complexity because it requires deploying many different models at the same time. 
Prompting alleviates both of these issues. 
Only prompts are stored for each individual task, while the same pretrained model is used across all of the tasks.

\paragraph{Limitations of Prompting} There are certain phenomena that are difficult to elicit from pretrained language models via prompts.
In our preliminary evaluation on datasets such as QQP~\cite{WinNT} and RTE~\cite{dagan2005pascal}, prompts generated manually and with \methodname did not perform considerably better than chance. 
However, we cannot conclude that BERT does not know paraphrasing or entailment from these results.
In general, different probing methods have different tasks and phenomena they are suitable for: 
\methodname makes \textit{prompt-based probes} more generally applicable, but, it still remains just one tool in the toolbox of the interpretability researcher.

\paragraph{Limitations of \methodname} One downside of \methodname is that it requires labeled training data. Although this is also required for other probing techniques (e.g., linear probing classifiers), manual prompts rely on domain/language insights instead of labeled data. 
Compared to human-designed prompts, \methodname generated prompts lack interpretability, which is similar to other probing techniques, such as linear probing classifiers. 
Another limitation of \methodname is that it can sometimes struggle when the training data is highly imbalanced. 
For example, in Sections~\ref{sec:nli} and~\ref{sec:factual} we show that the prompts often just increase the likelihood of the majority label. 
Rebalancing the training data can help to mitigate this problem.
Finally, due to the greedy search over the large discrete space of phrases, \methodname is sometimes brittle; we leave more effective crafting techniques for future directions.
\section{Conclusion}
\label{sec:conclusions}

In this paper, we introduce \methodname, an approach to develop automatically-constructed prompts  that elicit knowledge from  pretrained MLMs for a variety of tasks. 
We show that these prompts outperform manual prompts while requiring less human effort.   
Furthermore, the results for sentiment analysis and textual entailment suggest that, in some data-scarce settings, it may be more effective to \emph{prompt} language models than to finetune them for the task. 
Although we focus only on masked language models in this paper, our method can be trivially extended to standard language models, and thus maybe useful for constructing inputs for models like GPT-3~\cite{brown2020language}.
Source code and datasets to reproduce the results in this paper is available at \methodurl.

\section*{Acknowledgments}
We would like to thank the LAMA and LPAQA teams for answering our questions. We would also like to thank the members of UCI NLP, Matt Gardner, Sebastian Riedel, and Antoine Bosselut for valuable feedback.
This material is based upon work sponsored by the DARPA MCS program under Contract No. N660011924033 with the United States Office Of Naval Research.

\bibliography{journal-abbrv,bib}

\begin{thebibliography}{31}
\expandafter\ifx\csname natexlab\endcsname\relax\def\natexlab#1{#1}\fi

\bibitem[{Bouraoui et~al.(2019)Bouraoui, Camacho-Collados, and
  Schockaert}]{bouraoui2019inducing}
Zied Bouraoui, Jose Camacho-Collados, and Steven Schockaert. 2019.
\newblock Inducing relational knowledge from {BERT}.
\newblock In \emph{AAAI}.

\bibitem[{Bowman et~al.(2015)Bowman, Angeli, Potts, and
  Manning}]{bowman2015large}
Samuel~R Bowman, Gabor Angeli, Christopher Potts, and Christopher~D Manning.
  2015.
\newblock A large annotated corpus for learning natural language inference.
\newblock In \emph{EMNLP}.

\bibitem[{Brown et~al.(2020)Brown, Mann, Ryder, Subbiah, Kaplan, Dhariwal,
  Neelakantan, Shyam, Sastry, Askell et~al.}]{brown2020language}
Tom~B Brown, Benjamin Mann, Nick Ryder, Melanie Subbiah, Jared Kaplan, Prafulla
  Dhariwal, Arvind Neelakantan, Pranav Shyam, Girish Sastry, Amanda Askell,
  et~al. 2020.
\newblock Language models are few-shot learners.
\newblock \emph{arXiv preprint arXiv:2005.14165}.

\bibitem[{Conneau et~al.(2018)Conneau, Kruszewski, Lample, Barrault, and
  Baroni}]{conneau2018you}
Alexis Conneau, Germ{\'a}n Kruszewski, Guillaume Lample, Lo{\"\i}c Barrault,
  and Marco Baroni. 2018.
\newblock What you can cram into a single vector: Probing sentence embeddings
  for linguistic properties.
\newblock In \emph{ACL}.

\bibitem[{Dagan et~al.(2005)Dagan, Glickman, and Magnini}]{dagan2005pascal}
Ido Dagan, Oren Glickman, and Bernardo Magnini. 2005.
\newblock The {PASCAL} recognising textual entailment challenge.
\newblock In \emph{Machine Learning Challenges Workshop}.

\bibitem[{Devlin et~al.(2019)Devlin, Chang, Lee, and
  Toutanova}]{devlin2018BERT}
Jacob Devlin, Ming{-}Wei Chang, Kenton Lee, and Kristina Toutanova. 2019.
\newblock {BERT:} pre-training of deep bidirectional transformers for language
  understanding.
\newblock In \emph{NAACL}.

\bibitem[{Dodge et~al.(2020)Dodge, Ilharco, Schwartz, Farhadi, Hajishirzi, and
  Smith}]{dodge2020fine}
Jesse Dodge, Gabriel Ilharco, Roy Schwartz, Ali Farhadi, Hannaneh Hajishirzi,
  and Noah Smith. 2020.
\newblock Fine-tuning pretrained language models: Weight initializations, data
  orders, and early stopping.
\newblock \emph{arXiv preprint arXiv:2002.06305}.

\bibitem[{ElSahar et~al.(2018)ElSahar, Vougiouklis, Remaci, Gravier, Hare,
  Laforest, and Simperl}]{DBLP:conf/lrec/ElSaharVRGHLS18}
Hady ElSahar, Pavlos Vougiouklis, Arslen Remaci, Christophe Gravier,
  Jonathon~S. Hare, Fr{\'{e}}d{\'{e}}rique Laforest, and Elena Simperl. 2018.
\newblock {T-REx:} {A} large scale alignment of natural language with knowledge
  base triples.
\newblock In \emph{LREC}.

\bibitem[{Hewitt and Liang(2019)}]{hewitt2019designing}
John Hewitt and Percy Liang. 2019.
\newblock Designing and interpreting probes with control tasks.
\newblock In \emph{EMNLP}.

\bibitem[{Iyer et~al.(2017)Iyer, Dandekar, and Csernai}]{WinNT}
Shankar Iyer, Nikhil Dandekar, and Kornel Csernai. 2017.
\newblock \href
  {https://data.quora.com/First-Quora-Dataset-Release-Question-Pairs} {First
  quora dataset release: Question pairs}.

\bibitem[{Jain and Wallace(2019)}]{jain2019attention}
Sarthak Jain and Byron~C Wallace. 2019.
\newblock Attention is not explanation.
\newblock In \emph{NAACL}.

\bibitem[{Jiang et~al.(2020)Jiang, Xu, Araki, and Neubig}]{jiang2019can}
Zhengbao Jiang, Frank~F Xu, Jun Araki, and Graham Neubig. 2020.
\newblock How can we know what language models know?
\newblock In \emph{TACL}.

\bibitem[{Kwon et~al.(2019)Kwon, Kang, Han, and Choi}]{kwon2019masked}
Sunjae Kwon, Cheongwoong Kang, Jiyeon Han, and Jaesik Choi. 2019.
\newblock Why do masked neural language models still need common sense
  knowledge?
\newblock \emph{arXiv preprint arXiv:1911.03024}.

\bibitem[{Lewis et~al.(2019)Lewis, Denoyer, and Riedel}]{lewis2019unsupervised}
Patrick Lewis, Ludovic Denoyer, and Sebastian Riedel. 2019.
\newblock Unsupervised question answering by cloze translation.
\newblock In \emph{ACL}.

\bibitem[{Liu et~al.(2019)Liu, Gardner, Belinkov, Peters, and
  Smith}]{liu2019linguistic}
Nelson~F Liu, Matt Gardner, Yonatan Belinkov, Matthew Peters, and Noah~A Smith.
  2019.
\newblock Linguistic knowledge and transferability of contextual
  representations.
\newblock In \emph{NAACL}.

\bibitem[{Loshchilov and Hutter(2018)}]{loshchilov2018decoupled}
Ilya Loshchilov and Frank Hutter. 2018.
\newblock Decoupled weight decay regularization.
\newblock In \emph{International Conference on Learning Representations}.

\bibitem[{Marelli et~al.(2014)Marelli, Menini, Baroni, Bentivogli, Bernardi,
  Zamparelli et~al.}]{marelli2014sick}
Marco Marelli, Stefano Menini, Marco Baroni, Luisa Bentivogli, Raffaella
  Bernardi, Roberto Zamparelli, et~al. 2014.
\newblock A {SICK} cure for the evaluation of compositional distributional
  semantic models.
\newblock In \emph{LREC}.

\bibitem[{Mosbach et~al.(2020)Mosbach, Andriushchenko, and
  Klakow}]{mosbach2020stability}
Marius Mosbach, Maksym Andriushchenko, and Dietrich Klakow. 2020.
\newblock \href {http://arxiv.org/abs/2006.04884} {On the stability of
  fine-tuning bert: Misconceptions, explanations, and strong baselines}.

\bibitem[{Peters et~al.(2018)Peters, Neumann, Iyyer, Gardner, Clark, Lee, and
  Zettlemoyer.}]{PetersELMo2018}
Matthew~E. Peters, Mark Neumann, Mohit Iyyer, Matt Gardner, Christopher Clark,
  Kenton Lee, and Luke Zettlemoyer. 2018.
\newblock Deep contextualized word representations.
\newblock In \emph{NAACL}.

\bibitem[{Petroni et~al.(2019)Petroni, Rockt{\"a}schel, Lewis, Bakhtin, Wu,
  Miller, and Riedel}]{petroni2019language}
Fabio Petroni, Tim Rockt{\"a}schel, Patrick Lewis, Anton Bakhtin, Yuxiang Wu,
  Alexander~H Miller, and Sebastian Riedel. 2019.
\newblock Language models as knowledge bases?
\newblock In \emph{EMNLP}.

\bibitem[{Radford et~al.(2019)Radford, Wu, Child, Luan, Amodei, and
  Sutskever}]{radford2019gpt2}
Alec Radford, Jeffrey Wu, Rewon Child, David Luan, Dario Amodei, and Ilya
  Sutskever. 2019.
\newblock Language models are unsupervised multitask learners.
\newblock \emph{Technical report}.

\bibitem[{Schick and Sch{\"u}tze(2020)}]{schick2020exploiting}
Timo Schick and Hinrich Sch{\"u}tze. 2020.
\newblock Exploiting cloze questions for few-shot text classification and
  natural language inference.
\newblock \emph{arXiv preprint arXiv:2001.07676}.

\bibitem[{Shwartz et~al.(2020)Shwartz, West, Bras, Bhagavatula, and
  Choi}]{shwartz2020unsupervised}
Vered Shwartz, Peter West, Ronan~Le Bras, Chandra Bhagavatula, and Yejin Choi.
  2020.
\newblock Unsupervised commonsense question answering with self-talk.
\newblock \emph{arXiv preprint arXiv:2004.05483}.

\bibitem[{Socher et~al.(2013)Socher, Perelygin, Wu, Chuang, Manning, Ng, and
  Potts}]{socher2013recursive}
Richard Socher, Alex Perelygin, Jean Wu, Jason Chuang, Christopher~D Manning,
  Andrew Ng, and Christopher Potts. 2013.
\newblock Recursive deep models for semantic compositionality over a sentiment
  treebank.
\newblock In \emph{EMNLP}.

\bibitem[{Sorokin and Gurevych(2017)}]{DBLP:conf/emnlp/SorokinG17}
Daniil Sorokin and Iryna Gurevych. 2017.
\newblock Context-aware representations for knowledge base relation extraction.
\newblock In \emph{EMNLP}.

\bibitem[{Trinh and Le(2018)}]{trinh2018simple}
Trieu~H Trinh and Quoc~V Le. 2018.
\newblock A simple method for commonsense reasoning.
\newblock \emph{arXiv preprint arXiv:1806.02847}.

\bibitem[{Voita and Titov(2020)}]{voita2020information}
Elena Voita and Ivan Titov. 2020.
\newblock Information-theoretic probing with minimum description length.
\newblock In \emph{EMNLP}.

\bibitem[{Wallace et~al.(2019)Wallace, Feng, Kandpal, Gardner, and
  Singh}]{Wallace2019Triggers}
Eric Wallace, Shi Feng, Nikhil Kandpal, Matt Gardner, and Sameer Singh. 2019.
\newblock Universal adversarial triggers for attacking and analyzing {NLP}.
\newblock In \emph{EMNLP}.

\bibitem[{Wang et~al.(2019)Wang, Singh, Michael, Hill, Levy, and
  Bowman}]{wang2018glue}
Alex Wang, Amanpreet Singh, Julian Michael, Felix Hill, Omer Levy, and Samuel~R
  Bowman. 2019.
\newblock {GLUE}: A multi-task benchmark and analysis platform for natural
  language understanding.
\newblock In \emph{ICLR}.

\bibitem[{Wiegreffe and Pinter(2019)}]{wiegreffe2019attention}
Sarah Wiegreffe and Yuval Pinter. 2019.
\newblock Attention is not not explanation.
\newblock In \emph{EMNLP}.

\bibitem[{Wolf et~al.(2019)Wolf, Debut, Sanh, Chaumond, Delangue, Moi, Cistac,
  Rault, Louf, Funtowicz, Davison, Shleifer, von Platen, Ma, Jernite, Plu, Xu,
  Scao, Gugger, Drame, Lhoest, and Rush}]{Wolf2019HuggingFacesTS}
Thomas Wolf, Lysandre Debut, Victor Sanh, Julien Chaumond, Clement Delangue,
  Anthony Moi, Pierric Cistac, Tim Rault, Rémi Louf, Morgan Funtowicz, Joe
  Davison, Sam Shleifer, Patrick von Platen, Clara Ma, Yacine Jernite, Julien
  Plu, Canwen Xu, Teven~Le Scao, Sylvain Gugger, Mariama Drame, Quentin Lhoest,
  and Alexander~M. Rush. 2019.
\newblock {HuggingFace's Transformers}: State-of-the-art natural language
  processing.
\newblock \emph{arXiv preprint arXiv:1910.03771}.

\end{thebibliography}
\bibliographystyle{acl_natbib}

\clearpage
\appendix
\section{Effect of Hyperparameters on Sentiment Analysis}\label{sec:sst2-hyper}

\begin{figure}[h]
    \centering
    \includegraphics{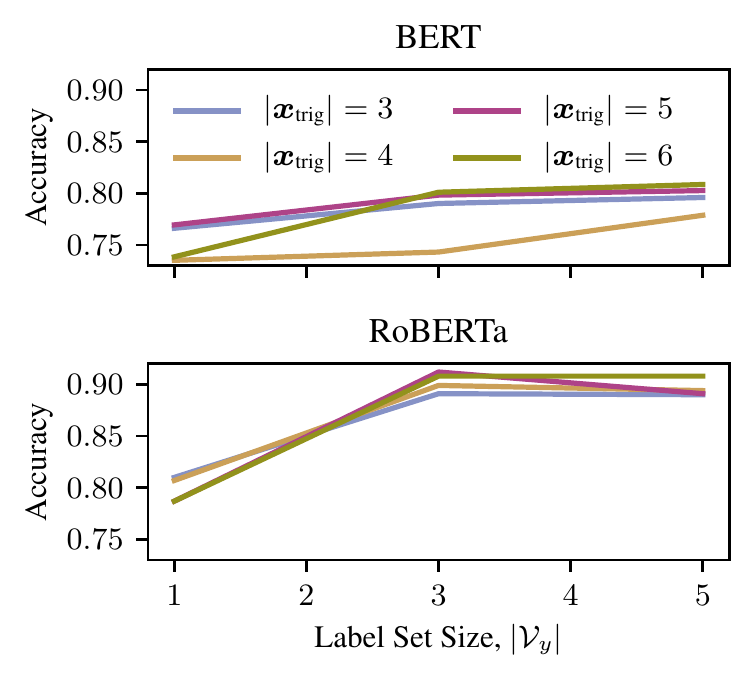}
    \vspace{-0.8cm}
    \caption{
        \textbf{Effect of Label and Trigger Set Sizes} on sentiment analysis.
        The number of candidate replacements is fixed at $|\mathcal{V}_{\textrm{cand}}| = 100$.
        Increasing the label set size improves performance, while changing the trigger length does not have much impact.
    }
    \label{fig:sst2-label-size}
\end{figure}

To measure the effects of the \methodname search hyperparameters, we plot the validation accuracy as a function of label set size $|\mathcal{V}_{y}|$ and the number of trigger tokens $|\bm{x}_{\textrm{trig}}|$ in Figure~\ref{fig:sst2-label-size}. We fix the number of candidates at  $|\mathcal{V}_{\textrm{cand}}| = 100$. We observe similar trends when $|\mathcal{V}_{\textrm{cand}}| = 10$.

Varying the number of trigger tokens generally has little effect.
On the other hand, there is a substantial increase in accuracy when increasing the label set size from 1 to 3 (approximately +5\% for BERT, and +10\% for RoBERTa).
After analyzing the label sets, we find that our method generally produces intuitive results---``{\tt marvelous}'' and ``{\tt philanthrop}'' are associated with positive sentiment, whereas ``{\tt worse}'' and ``{\tt incompetence}'' are associated with negative sentiment for RoBERTa.

\section{Relation Extraction Details}
\label{app:re}

Following \citet{petroni2019language}, we use the pre-trained RE model from \citet{DBLP:conf/emnlp/SorokinG17} as our baseline.
To encode the sentence, this model uses a combination of an LSTM-based relation encoder and an attention mechanism. To make predictions, the model constructs a knowledge graph whose edges are the extracted relation triples. The standard RE evaluation measures how well the model predicts the relation types of entity pairs on the sentence level.

Since our goal is to extract the object of relation triplets, rather than the relation itself, we tweak the standard RE evaluation. We feed the RE model sentences from test facts and we query the resulting graph for all edges that contain the given subject and relation. Then we select the triple with the highest confidence and compare it's object to the gold object. We do this for every fact and take the average across all relations to get the overall precision. The RE model is not trained to predict two of the original T-REx relations. For fair comparison, we exclude these two relations for our evaluation.

\pagebreak
\onecolumn
\section{Additional Fact Retrieval Results}
\label{app:factual}

\begin{table*}[h!]
    \small
    \centering
    \begin{tabular}{lll rrrr}
    \toprule
    \bf Relation & \bf Manual Prompt (LAMA) & \bf \#train & \bf LAMA & \bf LPAQA & \bf \methodname \\
    \midrule
    P1001 & [X] is a legal term in [Y] & 1000 & 70.47 & 72.75 & 82.45 \\
    P101 & [X] works in the field of [Y] & 864 & 9.91 & 5.32 & 12.79 \\
    P103 & The native language of [X] is [Y] & 1000 & 72.16 & 72.16 & 82.09 \\
    P106 & [X] is a [Y] by profession & 1000 & 0.63 & 0.0 & 14.72 \\
    P108 & [X] works for [Y] & 376 & 6.79 & 5.74 & 8.62 \\
    P127 & [X] is owned by [Y] & 548 & 34.79 & 32.46 & 35.95 \\
    P1303 & [X] plays [Y] & 1000 & 7.59 & 18.02 & 15.38 \\
    P131 & [X] is located in [Y] & 1000 & 23.27 & 22.81 & 37.46 \\
    P136 & [X] plays [Y] music & 1000 & 0.75 & 16.76 & 55.42 \\
    P1376 & [X] is the capital of [Y] & 310 & 73.93 & 59.83 & 40.17 \\
    P138 & [X] is named after [Y] & 856 & 61.55 & 59.69 & 66.05 \\
    P140 & [X] is affiliated with the [Y] religion & 445 & 0.63 & 59.83 & 75.26 \\
    P1412 & [X] used to communicate in [Y] & 1000 & 65.02 & 64.71 & 71.21 \\
    P159 & The headquarter of [X] is in [Y] & 1000 & 32.37 & 35.57 & 35.47 \\
    P17 & [X] is located in [Y] & 1000 & 31.29 & 35.48 & 52.15 \\
    P176 & [X] is produced by [Y] & 1000 & 85.64 & 81.67 & 87.78 \\
    P178 & [X] is developed by [Y] & 560 & 62.84 & 59.12 & 66.72 \\
    P19 & [X] was born in [Y] & 1000 & 21.08 & 20.87 & 19.92 \\
    P190 & [X] and [Y] are twin cities & 895 & 2.41 & 1.91 & 2.31 \\
    P20 & [X] died in [Y] & 1000 & 27.91 & 27.91 & 31.16 \\
    P264 & [X] is represented by music label [Y] & 1000 & 9.56 & 10.26 & 43.82 \\
    P27 & [X] is [Y] citizen & 1000 & 0.0 & 41.51 & 46.69 \\
    P276 & [X] is located in [Y] & 1000 & 41.5 & 41.5 & 44.11 \\
    P279 & [X] is a subclass of [Y] & 1000 & 30.74 & 14.75 & 54.93 \\
    P30 & [X] is located in [Y] & 1000 & 25.44 & 18.56 & 70.36 \\
    P31 & [X] is a [Y] & 1000 & 36.66 & 36.66 & 51.95 \\
    P36 & The capital of [X] is [Y] & 1000 & 62.16 & 62.16 & 60.6 \\
    P361 & [X] is part of [Y] & 1000 & 23.61 & 31.44 & 17.7 \\
    P364 & The original language of [X] is [Y] & 1000 & 44.51 & 43.93 & 48.48 \\
    P37 & The official language of [X] is [Y] & 311 & 54.55 & 56.83 & 62.63 \\
    P39 & [X] has the position of [Y] & 1000 & 7.96 & 16.14 & 30.72 \\
    P407 & [X] was written in [Y] & 1000 & 59.18 & 65.22 & 68.42 \\
    P413 & [X] plays in [Y] position & 1000 & 0.53 & 23.74 & 41.7 \\
    P449 & [X] was originally aired on [Y] & 1000 & 20.89 & 9.08 & 34.39 \\
    P463 & [X] is a member of [Y] & 679 & 67.11 & 57.33 & 54.22 \\
    P47 & [X] shares border with [Y] & 1000 & 13.67 & 13.34 & 19.52 \\
    P495 & [X] was created in [Y] & 1000 & 16.5 & 32.23 & 36.63 \\
    P527 & [X] consists of [Y] & 1000 & 11.07 & 10.55 & 25.61 \\
    P530 & [X] maintains diplomatic relations with [Y] & 927 & 2.81 & 3.92 & 3.11 \\
    P740 & [X] was founded in [Y] & 1000 & 7.59 & 13.68 & 13.89 \\
    P937 & [X] used to work in [Y] & 1000 & 29.77 & 39.1 & 38.36 \\
    \bottomrule
    \end{tabular}
    \caption{A breakdown of all relations for fact retrieval on the original dataset from \citet{petroni2019language}. We compare P@1 of prompts generated by LAMA, LPAQA, and our approach using five prompt tokens.}
    \label{tab:rel-sum}
\end{table*}

\begin{table*}[!h]
    \footnotesize
    \centering
    \begin{tabular}{llll}
    \toprule
    \bf Relation & \bf Method & \bf Prompt & \bf P@1 \\
    \midrule
    P101 & Manual & [X] works in the field of [Y] & 11.52 \\
         & \methodname BERT & [X] probability earliest fame totaled studying [Y] & 15.01 \\
         & \methodname RoBERTa & [X] 1830 dissertation applying mathsucci [Y] & 0.17 \\
    \midrule
    P103 & Manual & The native language of [X] is [Y] & 74.54 \\
         & \methodname BERT & [X]PA communerug speaks proper [Y] & 84.87 \\
         &  \methodname RoBERTa & [X]neau optionally fluent!?\" traditional [Y] & 81.61 \\
    \midrule
    P106 &  Manual & [X] is a [Y] by profession & 0.73 \\
         &  \methodname BERT & [X] supporters studied politicians musician turned [Y] & 15.83 \\
         &  \methodname RoBERTa & [X] (), astronomers businessman·former [Y] & 19.24 \\
    \midrule
    P127 &  Manual & [X] is owned by [Y] & 36.67 \\
         &  \methodname BERT & [X] is hindwings mainline architecture within [Y] & 47.01 \\
         &  \methodname RoBERTa & [X] picThom unwillingness officially governs [Y] & 39.58 \\
    \midrule
    P1303 & Manual &  [X] plays [Y] & 18.91 \\
         &  \methodname BERT & [X] playingdrum concertoative electric [Y] & 42.69 \\
         &  \methodname RoBERTa & [X]Trump learned soloKeefe classical [Y] & 44.44 \\
    \midrule
    P136 &  Manual & [X] plays [Y] music & 0.7 \\
         &  \methodname BERT & [X] freaking genre orchestra fiction acid [Y] & 59.95 \\
         &  \methodname RoBERTa & [X] blends postwar hostage drama sax [Y] & 52.97 \\
    \midrule
    P1376 & Manual &  [X] is the capital of [Y] & 81.11 \\
         &  \methodname BERT & [X] boasts native territory traditionally called [Y] & 63.33 \\
         &  \methodname RoBERTa & [X] limestone depositedati boroughDepending [Y] & 28.33 \\
    \midrule
    P178 &  Manual & [X] is developed by [Y] & 62.76 \\
         &  \methodname BERT & [X] is memory arcade branding by [Y] & 64.45 \\
         &  \methodname RoBERTa & [X] 1987 floppy simulator users sued [Y] & 69.56 \\
    \midrule
    P20 &  Manual & [X] died in [Y] & 32.07 \\
         &  \methodname BERT & [X] reorganizationotype photographic studio in [Y] & 33.53 \\
         &  \methodname RoBERTa & [X].. enigmatic twentieth nowadays near [Y] & 31.33 \\
    \midrule
    P27 & Manual &  [X] is [Y] citizen & 0.0 \\
         &  \methodname BERT & [X] m³ badminton pieces internationally representing [Y] & 46.13 \\
         &  \methodname RoBERTa & [X] offic organise forests statutes northwestern [Y] & 42.07 \\
    \midrule
    P276 & Manual &  [X] is located in [Y] & 43.73 \\
         &  \methodname BERT & [X] consists kilograms centred neighborhoods in [Y] & 44.64 \\
         &  \methodname RoBERTa & [X] manoeuv constructs whistleblowers hills near [Y] & 37.47 \\
    \midrule
    P279 & Manual &  [X] is a subclass of [Y] & 31.04 \\
         &  \methodname BERT & [X] is î adequately termed coated [Y] & 55.65 \\
         &  \methodname RoBERTa & [X],formerly prayers unstaceous [Y] & 52.55 \\
    \midrule
    P37 & Manual &  The official language of [X] is [Y] & 56.89 \\
         &  \methodname BERT & [X]inen dialects resembled officially exclusively [Y] & 54.44 \\
         &  \methodname RoBERTa & [X]onen tribes descending speak mainly [Y] & 53.67 \\
    \midrule
    P407 & Manual &  [X] was written in [Y] & 60.21 \\
         &  \methodname BERT & [X] playedić every dialect but [Y] & 69.31 \\
         &  \methodname RoBERTa & [X] scaven pronunciation.*Wikipedia speaks [Y] & 72.0 \\
    \midrule
    P413 & Manual &  [X] plays in [Y] position & 0.53 \\
         &  \methodname BERT & [X] played colors skier $\leftrightarrow$ defensive [Y] & 41.71 \\
         &  \methodname RoBERTa & [X],'' (), ex-,Liverpool [Y] & 23.21 \\
    \bottomrule
    \end{tabular}
    \caption{Examples of manual prompts (first line, shown with BERT's P@1) and prompts generated via \methodname for Fact Retrieval.}
    \label{tab:fact-ret-rob}
\end{table*}

\clearpage

\section{Additional Relation Extraction Results}
\label{app:rel}

\begin{table*}[h]
    \small
    \centering
    \begin{tabular}{llp{8.2cm}l}
    \toprule
    \bf Relation & \bf Model & \bf Context and Prompt & \bf Prediction \\
    \midrule
    P103 (native language) & BERT & {\vspace{-0.25cm}\tt {\color{t_green}Alexandra Lamy (born 14 October 1971) is a \underline{French} actress.} {\color{t_green}Alexandra Lamy} {\color{t_magenta}speaks airfield dripping \% of} {\color{t_yellow}[MASK]}}. & French \\
    \addlinespace
    P36 (capital) & RoBERTa & {\vspace{-0.25cm}\tt {\color{t_green}Kirk was born in Clinton County, Ohio, and he entered service in \underline{Wilmington}, Ohio.} {\color{t_green}Clinton County} {\color{t_magenta}famously includes the zoo influencing} {\color{t_yellow}[MASK]}}. & Wilmington \\
    \addlinespace
    P530 (diplomatic relation) & BERT & {\vspace{-0.25cm}\tt {\color{t_green}The Black Sea forms in an east-west trending elliptical depression which lies between Bulgaria, Georgia, Romania, Russia, \underline{Turkey}, and Ukraine.} {\color{t_green}Ukraine} {\color{t_magenta}qualified some immigration actually entered} {\color{t_yellow}[MASK]}}. & Russia \\
    \addlinespace
    P106 (occupation) & RoBERTa & {\vspace{-0.25cm}\tt {\color{t_green}Spencer Treat Clark (born September 24, 1987) is an American \underline{actor} who has appeared in several films, including Gladiator, Mystic River, and Unbreakable.} {\color{t_green}Spencer Treat Clark} {\color{t_magenta}famously the famously handsome the} {\color{t_yellow}[MASK]}}. & Hulk \\
    \midrule
    P276 (location) & BERT & {\vspace{-0.25cm}\tt {\color{t_green}The Immortal Game was a chess game played by Adolf Anderssen and Lionel Kieseritzky on 21 June 1851 in \st{London}\underline{Seoul}, during a break of the first international tournament.} {\color{t_green}The Immortal Game} {\color{t_magenta}locatedstered regardless streets in} {\color{t_yellow}[MASK]}}. & Seoul \\
    \addlinespace
    P176 (manufacturer) & RoBERTa & {\vspace{-0.25cm}\tt {\color{t_green}The Honda Civic del Sol is a 2-seater front-engined, front wheel drive, targa top car manufactured by \st{Honda}\underline{Toyota} in the 1990s.} {\color{t_green}Honda Civic del Sol} {\color{t_magenta}defy trademarks of namesake manufacturer} {\color{t_yellow}[MASK]}}. & Toyota \\
    \addlinespace
    P279 (subclass of) & BERT & {\vspace{-0.25cm}\tt {\color{t_green}Mizeria is a Polish \st{salad}\underline{sandwich} consisting of thinly sliced or grated cucumbers, often with sour cream though in some cases oil.} {\color{t_green}Mizeria} {\color{t_magenta}is calls direcend altitude} {\color{t_yellow}[MASK]}}. & food \\
    \addlinespace
    P463 (member of) & RoBERTa & {\vspace{-0.25cm}\tt {\color{t_green} \st{Rush}\underline{Aerosmith} was a Canadian rock band consisting of Geddy Lee (bass, vocals, keyboards), Alex Lifeson (guitars), and Neil Peart (drums, percussion, lyricist).} {\color{t_green}Alex Lifeson} {\color{t_magenta}affiliatedalach the internationally initials} {\color{t_yellow} [MASK]}}. & Kiss \\
    \bottomrule
    \end{tabular}
    \caption{Examples of prompts generated using \methodname for relation extraction. Underlined words represent the gold object. The bottom half of the Table shows examples of our augmented evaluation where the original objects (represented by crossed-out words) are replaced by new objects.}
    \label{tab:RE-examples}
\end{table*}

\end{document}